\pdfoutput=1

\documentclass[11pt]{article}

\usepackage{EMNLP2023}

\usepackage{times}
\usepackage{latexsym}

\usepackage[T1]{fontenc}

\usepackage[utf8]{inputenc}
\usepackage{tikz}

\usepackage{microtype}
\usepackage{graphicx}
\usepackage{inconsolata}
\usepackage{comment}
\usepackage{xspace}
\usepackage{multirow}
\usepackage{multicol}
\usepackage{graphicx}
\usepackage{booktabs}
\usepackage{todonotes}
\usepackage{tablefootnote}
\usepackage{soul}
\usepackage{subfigure}
\usepackage{listings}
\newcommand{\system}{\textsc{QACheck}\xspace}
\usepackage{listings}
\definecolor{codegreen}{rgb}{0,0.6,0}
\definecolor{codegray}{rgb}{0.5,0.5,0.5}
\definecolor{codepurple}{rgb}{0.58,0,0.82}

\lstdefinestyle{mystyle}{
  frame=single,
  basicstyle=\ttfamily\footnotesize,
  backgroundcolor=\color{backcolour}, commentstyle=\color{codegreen},
  commentstyle=\color{darkgreen}\slshape,
  keywordstyle=\color{blue},
  stringstyle=\color{darkred},
  numberstyle=\tiny\color{codegray},
  emphstyle=\color{pink}\underbar,
  morekeywords={Verify, Question},
  escapeinside={(*@}{@*)},
  breakatwhitespace=false,         
  breaklines=true,                 
  captionpos=b,                    
  keepspaces=true,                    
  numbersep=5pt,                  
  showspaces=false,                
  showstringspaces=false,
  showtabs=false,                  
  tabsize=2
}
\newcommand{\authorspace}{\hspace{0.3cm}}

\title{\system: A Demonstration System\\ for Question-Guided Multi-Hop Fact-Checking}

\author{
    \bf Liangming Pan$^{1,2}$\authorspace
  \bf Xinyuan Lu$^3$ \authorspace
  \bf Min-Yen Kan$^3$ \authorspace
  \bf Preslav Nakov$^1$ \authorspace
  \\
 $^1$MBZUAI \authorspace
 $^2$University of California, Santa Barbara \authorspace 
  $^3$ National University of Singapore \\
 \\
  {\tt liangmingpan@ucsb.edu} \authorspace \authorspace
  {\tt luxinyuan@u.nus.edu} \authorspace \authorspace \\
  {\tt kanmy@comp.nus.edu.sg} \authorspace \authorspace
    {\tt preslav.nakov@mbzuai.ac.ae} 
}

\begin{document}
\maketitle
\begin{abstract}
Fact-checking real-world claims often requires complex, multi-step reasoning due to the absence of direct evidence to support or refute them. However, existing fact-checking systems often lack transparency in their decision-making, making it challenging for users to comprehend their reasoning process. To address this, we propose the \textit{Question-guided Multi-hop Fact-Checking} (\system) system, which guides the model's reasoning process by asking a series of questions critical for verifying a claim. \system has five key modules: a claim verifier, a question generator, a question-answering module, a QA validator, and a reasoner. Users can input a claim into \system, which then predicts its veracity and provides a comprehensive report detailing its reasoning process, guided by a sequence of (question, answer) pairs. \system \footnote{\system is public available at \url{https://github.com/XinyuanLu00/QACheck}. A recorded video is at \url{https://www.youtube.com/watch?v=ju8kxSldM64}} also provides the source of evidence supporting each question, fostering a transparent, explainable, and user-friendly fact-checking process. 

\end{abstract}

\section{Introduction}
In our age characterized by large amounts of both true and false information, fact-checking is not only crucial for counteracting misinformation but also plays a vital role in fostering trust in AI systems. However, the process of validating real-world claims is rarely straightforward. Unlike the simplicity of supporting or refuting a claim with a single piece of direct evidence, real-world claims often resemble multi-layered puzzles that require complex and multi-step reasoning to solve~\cite{Jiang2020HoVerAD,FANG,aly-vlachos-2022-natural,Chen2022GeneratingLA,pan-etal-2023-fact}. 

\begin{figure}[!t]
\centering
\includegraphics[width=7.5cm]{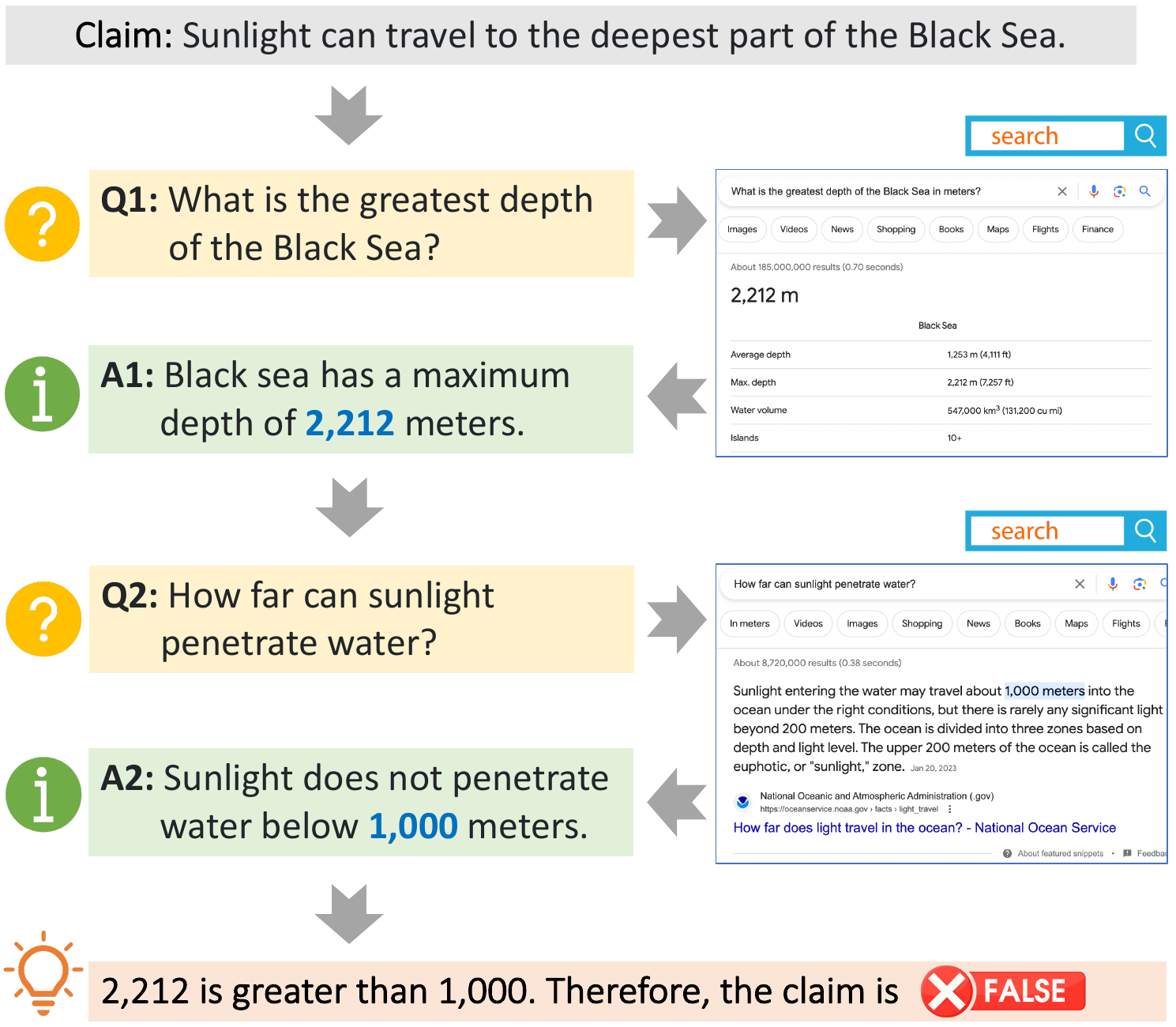}
\caption{An example of \textit{question-guided} reasoning for fact-checking complex real-world claims. }
\label{fig:example}
\vspace{-0.3cm}
\end{figure}

As an example, to verify the claim \textit{``Sunlight can reach the deepest part of the Black Sea.''}, it may be challenging to find direct evidence on the web that refutes or supports this claim. Instead, a human fact-checker needs to decompose the claim, gather multiple pieces of evidence, and perform step-by-step reasoning~\cite{pan-etal-2023-fact}. This reasoning process can be formulated as \textit{question-guided reasoning}, where the verification of the claim is guided by asking and answering a series of relevant questions, as shown in Figure~\ref{fig:example}. In this example, we sequentially raise two questions: \textit{``What is the greatest depth of the Black Sea?''} and \textit{``How far can sunlight penetrate water?''}. After independently answering these two questions by gathering relevant information from the Web, we can assert that the initial claim is \textit{false} with simple reasoning. 


While several models~\cite{DBLP:conf/acl/LiuXSL20,DBLP:conf/acl/ZhongXTXDZWY20,aly-vlachos-2022-natural} have been proposed to facilitate multi-step reasoning in fact-checking, they generally lack transparency in their reasoning processes. These models simply take a claim as input, then output a veracity label without an explicit explanation. Recent attempts, such as \textit{Quin+}~\cite{DBLP:conf/naacl/SamarinasHL21} and \textit{WhatTheWikiFact}~\cite{chernyavskiy2021whatthewikifact}, have aimed to develop more explainable fact-checking systems, by searching and visualizing the supporting evidence for a given claim. However, these systems primarily validate a claim from a \textit{single} document, and do not provide a detailed, step-by-step visualization of the reasoning process as shown in Figure~\ref{fig:example}. 

\begin{figure}[!t]
    \centering
    \includegraphics[width=7.5cm]{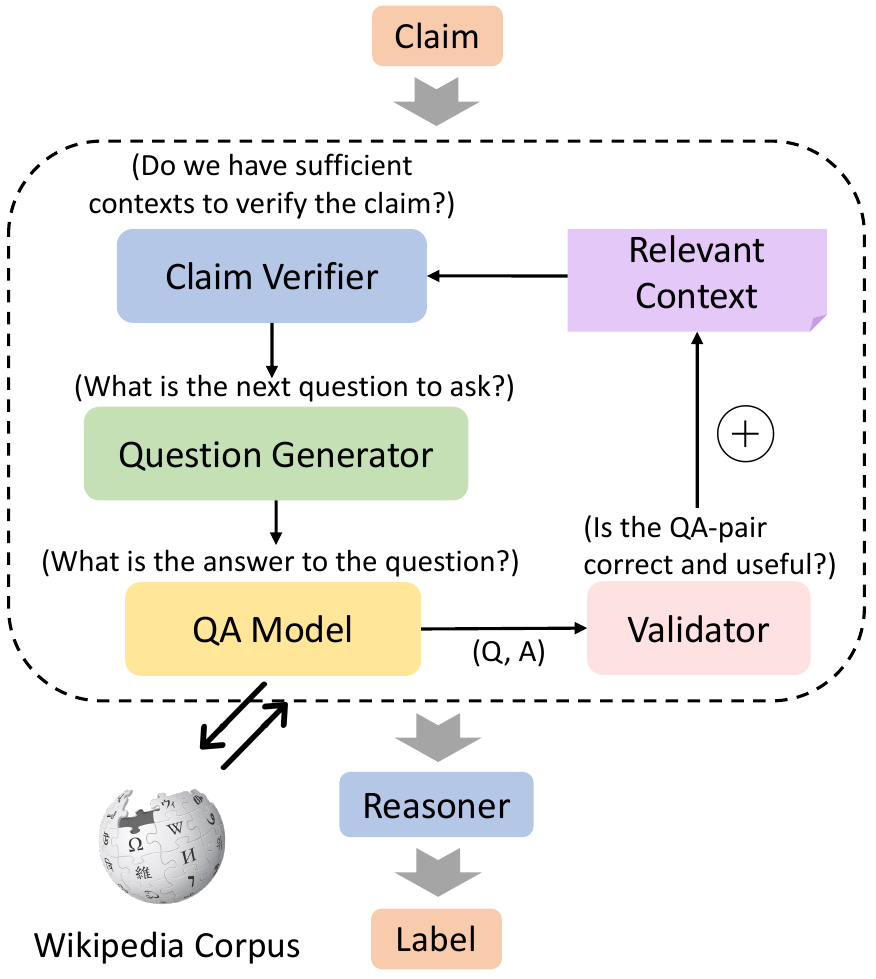}
    \caption{The architecture of our \system system.}
    \label{fig:overall architecture}
\end{figure}

We introduce the \textit{Question-guided Multi-hop Fact-Checking} (\system) system, which 
addresses the aforementioned issues by generating multi-step explanations via question-guided reasoning. To facilitate an explainable reasoning process, \system manages the reasoning process by guiding the model to self-generate a series of questions vital for claim verification. Our system, as depicted in Figure~\ref{fig:overall architecture}, is composed of five modules: 1) a \textit{claim verifier} that assesses whether sufficient information has been gathered to verify the claim, 2) a \textit{question generator} to generate the next relevant question, 3) a \textit{question-answering} module to answer the raised question, 4) a \textit{QA validator} to evaluate the usefulness of the generated (Q, A) pair, and 5) a \textit{reasoner} to output the final veracity label based on all collected contexts. 

\system offers enough adaptability, allowing users to customize the design of each module by integrating with different models. 
For example, we provide three alternative implementations for the QA component: the retriever--reader model, the FLAN-T5 model, and the GPT3-based reciter--reader model. Furthermore, we offer a user-friendly interface for users to fact-check any input claim and visualize its detailed question-guided reasoning process. The screenshot of our user interface is shown in Figure~\ref{fig:interface}. We will discuss the implementation details of the system modules in Section~\ref{sec:system_overview} and some evaluation results in Section~\ref{sec:evaluation}. Finally, we present the details of the user interface in Section~\ref{sec:interface}.  and conclude and discuss future work in Section~\ref{sec:conclusion}. 


\section{Related Work}
\label{sec:related_work}

\paragraph{Fact-Checking Systems.} The recent surge in automated fact-checking research aims to mitigate the spread of misinformation. Various fact-checking systems, for example, \textsc{Tanbih}\footnote{\url{https://www.tanbih.org/about}}~\cite{zhang-etal-2019-tanbih}, \textsc{PRTA}\footnote{\url{https://propaganda.qcri.org/}}~\cite{da-san-martino-etal-2020-prta}, and \textsc{WhatTheWikiFact}\footnote{\url{https://www.tanbih.org/whatthewikifact}}
~\cite{chernyavskiy2021whatthewikifact} predominantly originating from Wikipedia and claims within political or scientific domains, have facilitated this endeavor. However, the majority of these systems limit the validation or refutation of a claim to a single document, indicating a gap in systems for multi-step reasoning~\cite{pan-etal-2023-fact}. The system most similar to ours is \textit{Quin+}~\cite{DBLP:conf/naacl/SamarinasHL21}, which demonstrates evidence retrieval in a single step. In contrast, our \system shows a question-led multi-step reasoning process with explanations and retrieved evidence for each reasoning step. In summary, our system 1) supports fact-checking real-world claims that require multi-step reasoning, and 2) enhances transparency and helps users have a clear understanding of the reasoning process. 


\paragraph{Explanation Generation.}
Simply predicting a veracity label to the claim is not persuasive, and can even enhance mistaken beliefs~\cite{guo-etal-2022-survey}. Hence, it is necessary for automated fact-checking methods to provide explanations to support model predictions. Traditional approaches have utilized attention weights, logic,
or summary generation to provide post-hoc explanations for model predictions~\cite{DBLP:conf/acl/LuL20,DBLP:conf/tto/AhmadiLPS19,DBLP:conf/emnlp/KotonyaT20,DBLP:journals/information/JollyAA22, xing-etal-2022-automatic}. In contrast, our approach employs \textit{question--answer} pair based explanations, offering more human-like and natural explanations.

\section{System Architecture}
\label{sec:system_overview}

Figure~\ref{fig:overall architecture} shows the general architecture of our system, comprised of five principal modules: a Claim Verifier $\mathcal{D}$, a Question Generator $\mathcal{Q}$, a Question-Answering Model $\mathcal{A}$, a Validator $\mathcal{V}$, and a Reasoner $\mathcal{R}$. 
We first initialize an empty context $\mathcal{C} = \emptyset$. Upon the receipt of a new input claim $c$, the system first utilizes the \textit{claim verifier} to determine the sufficiency of the existing context to validate the claim, \textit{i.e.}, $\mathcal{D}(c, \mathcal{C}) \rightarrow \{\texttt{True}, \texttt{False}\}$. If the output is \texttt{False}, the \textit{question generator} learns to generate the next question that is necessary for verifying the claim, \textit{i.e.}, $\mathcal{Q}(c, \mathcal{C}) \rightarrow q$. The \textit{question-answering} model is then applied to answer the question and provide the supported evidence, \textit{i.e.}, $\mathcal{A}(q) \rightarrow a, e$, where $a$ is the predicted answer, and $e$ is the retrieved evidence that supports the answer. Afterward, the \textit{validator} is used to validate the usefulness of the newly-generated (Q, A) pair based on the existing context and the claim, \textit{i.e.}, $\mathcal{V}(c, \{q, a\}, \mathcal{C}) \rightarrow \{\texttt{True}, \texttt{False}\}$. If the output is \texttt{True}, the $(q, a)$ pair is added into the context $C$. Otherwise, the question generator is asked to generate another question. We repeat this process of calling $\mathcal{D} \rightarrow \mathcal{Q} \rightarrow \mathcal{A} \rightarrow \mathcal{V}$ until the claim verifier returns a \texttt{True} indicating that the current context $C$ contains sufficient information to verify the claim $c$. In this case, the \textit{reasoner} module is called to utilize the stored relevant context to justify the veracity of the claim and outputs the final label, \textit{i.e.}, $\mathcal{R}(c, \mathcal{C}) \rightarrow \{\texttt{Supported}, \texttt{Refuted}\}$. The subsequent sections provide a comprehensive description of the five key modules in \system. 

\subsection{Claim Verifier}
\label{sec: claim verifier}
The claim verifier is a central component of \system, with the specific role of determining if the current context information is sufficient to verify the claim. This module is to ensure that the system can efficiently complete the claim verification process without redundant reasoning. 
We build the claim verifier based on InstructGPT~\cite{DBLP:journals/corr/abs-2203-02155}, utilizing its powerful \textit{in-context learning} ability. Recent large language models such as InstructGPT~\cite{DBLP:journals/corr/abs-2203-02155} and GPT-4~\cite{DBLP:GPT4} have demonstrated strong few-shot generalization ability via \textit{in-context learning}, in which the model can efficiently learn a task when prompted with the instruction of the task together with a small number of demonstrations. We take advantage of InstructGPT's in-context learning ability to implement the claim verifier. 
We prompt InstructGPT with ten distinct in-context examples as detailed in Appendix~\ref{subsec:claim_verifier}, where each example consists of a claim and relevant question--answer pairs. We then prompt the model with the claim, the context, and the following instruction: 
\begin{quote}
\texttt{Claim = \colorbox{gray}{\color{white}{CLAIM}} \\
We already know the following:} \\
\texttt{\colorbox{gray}{\color{white}{CONTEXT}} \\
Can we know whether the claim is true or false now?
Yes or no?}
\end{quote}
\noindent If the response is \textit{`no'}, we proceed to the question generator module. Conversely, if the response is \textit{`yes'}, the process jumps to call the reasoner module. 

\subsection{Question Generator}
\label{sec: question generator}
The question generator module is called when the initial claim lacks the necessary context for verification. This module aims to generate the next relevant question needed for verifying the claim. Similar to the claim verifier, we also leverage InstructGPT for in-context learning. We use slightly different prompts for generating the initial question and the follow-up questions. The detailed prompts are in Appendix~\ref{subsec:question_generation}. For the \textit{initial} question generation, the instruction is: 
\begin{quote}
\texttt{Claim = \colorbox{gray}{\color{white}{CLAIM}} \\
To verify the above claim, we can first ask a simple question:}
\end{quote}
\noindent For \textit{follow-up} questions, the instruction is:
\begin{quote}
\texttt{Claim = \colorbox{gray}{\color{white}{CLAIM}} \\
We already know the following:} \\
\texttt{\colorbox{gray}{\color{white}{CONTEXT}} \\
To verify the claim, what is the next question we need to know the answer to?}
\end{quote}

\subsection{Question Answering Model}
\label{sec: qa model}
\begin{figure}[!t]
    \centering
    \includegraphics[width=7.5cm]{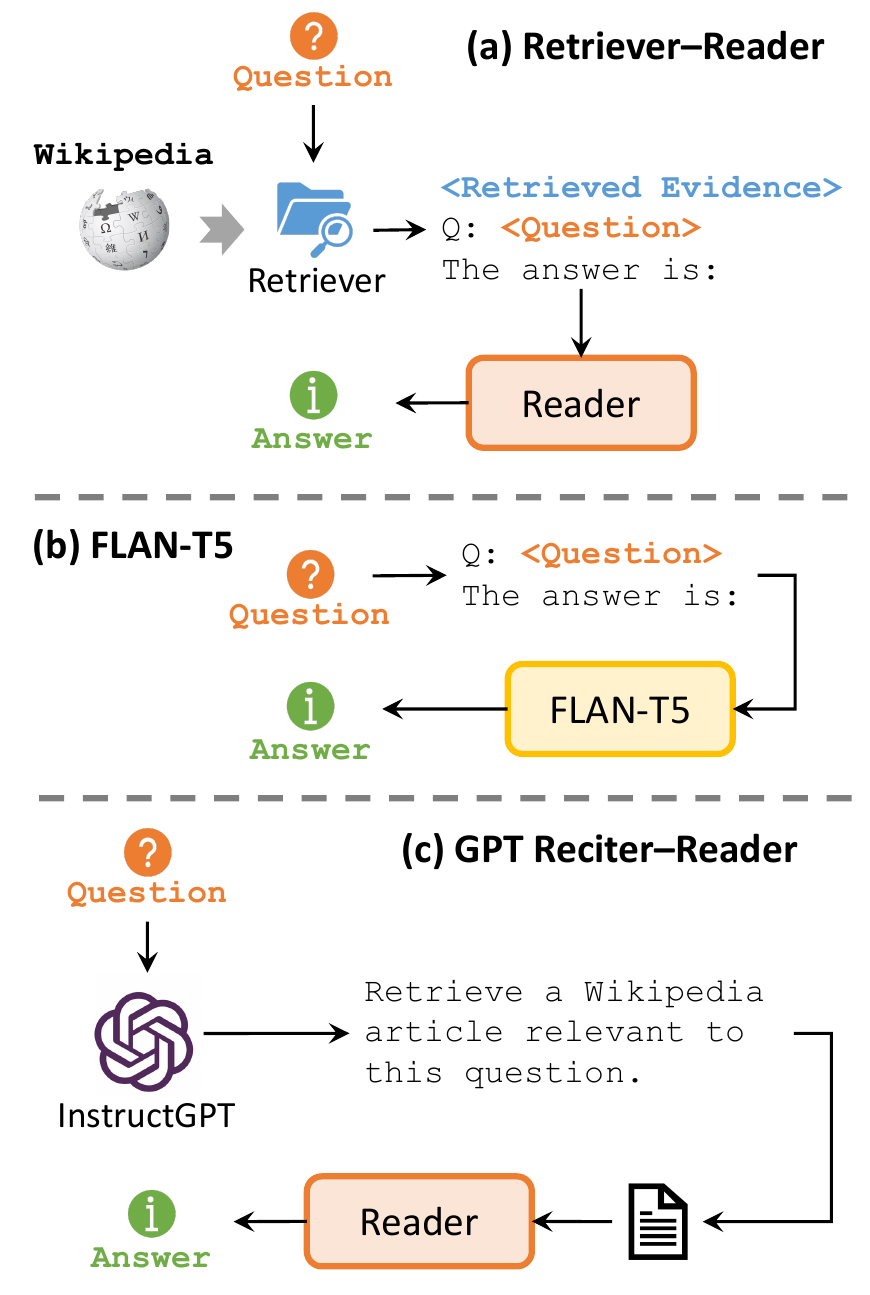}
    \caption{Illustrations of the three different implementations of the Question Answering module in \system. }
    \label{fig:qamethod}
\end{figure}

After generating a question, the Question Answering (QA) module retrieves corresponding evidence and provides an answer as the output. The system's reliability largely depends on the accuracy of the QA module's responses. 
Understanding the need for different QA methods in various fact-checking scenarios, we introduce three different implementations for the QA module, as shown in Figure~\ref{fig:qamethod}.

\paragraph{Retriever--Reader.} We first integrate the well-known \textit{retriever--reader} framework, a prevalent QA paradigm originally introduced by~\citet{DBLP:conf/acl/ChenFWB17}. In this framework, a \textit{retriever} first retrieves relevant documents from a large evidence corpus, and then a \textit{reader} predicts an answer conditioned on the retrieved documents. For the evidence corpus, we use the Wikipedia dump provided by the Knowledge-Intensive Language Tasks (KILT) benchmark~\cite{DBLP:conf/naacl/PetroniPFLYCTJK21}, in which the Wikipedia articles have been pre-processed and separated into paragraphs. For the retriever, we apply the widely-used sparse retrieval based on BM25~\cite{DBLP:BM25}, implemented with the Pyserini toolkit~\cite{DBLP:Pyserini}. For the reader, we use the \textit{RoBERTa-large}~\cite{DBLP:RoBerta} model fine-tuned on the SQuAD dataset~\cite{DBLP:SQuAD}, using the implementation from \textit{PrimeQA}\footnote{\url{https://github.com/primeqa/primeqa}}~\cite{DBLP:PrimeQA}. 

\paragraph{FLAN-T5.} 
While effective, the retriever--reader framework is constrained by its reliance on the evidence corpus. In scenarios where a user's claim is outside the scope of Wikipedia, the system might fail to produce a credible response. To enhance flexibility, we also incorporate the \textit{FLAN-T5} model~\cite{DBLP:journals/corr/abs-2210-11416}, a Seq2Seq model pretrained on more than 1.8K tasks with instruction tuning. It directly takes the question as input and then generates the answer and the evidence, based on the model's parametric knowledge. 

\paragraph{GPT Reciter--Reader.} 
Recent studies~\cite{sun2023recitationaugmented,yu2023generate} have demonstrated the great potential of the GPT series, such as InstructGPT~\cite{DBLP:journals/corr/abs-2203-02155} and GPT-4~\cite{DBLP:GPT4}, to function as robust knowledge repositories. The knowledge can be retrieved by properly prompting the model. Drawing from this insight, we introduce the \textit{GPT Reciter--Reader} approach. Given a question, we prompt the InstructGPT to \textit{``recite''} the knowledge stored within it, and InstructGPT responds with relevant evidence. The evidence is then fed into a \textit{reader} model to produce the corresponding answer. 
While this method, like FLAN-T5, does not rely on a specific corpus, it stands out by using InstructGPT. This offers a more dependable parametric knowledge base than FLAN-T5.

\vspace{0.1cm}

The above three methods provide a flexible and robust QA module, allowing for switching between the methods as required, depending on the claim being verified and the available contextual information. In the following, we use GPT Reciter--Reader as the default implementation for our QA module. 


\begin{figure*}[!t]
    \centering
    \includegraphics[width=16cm]{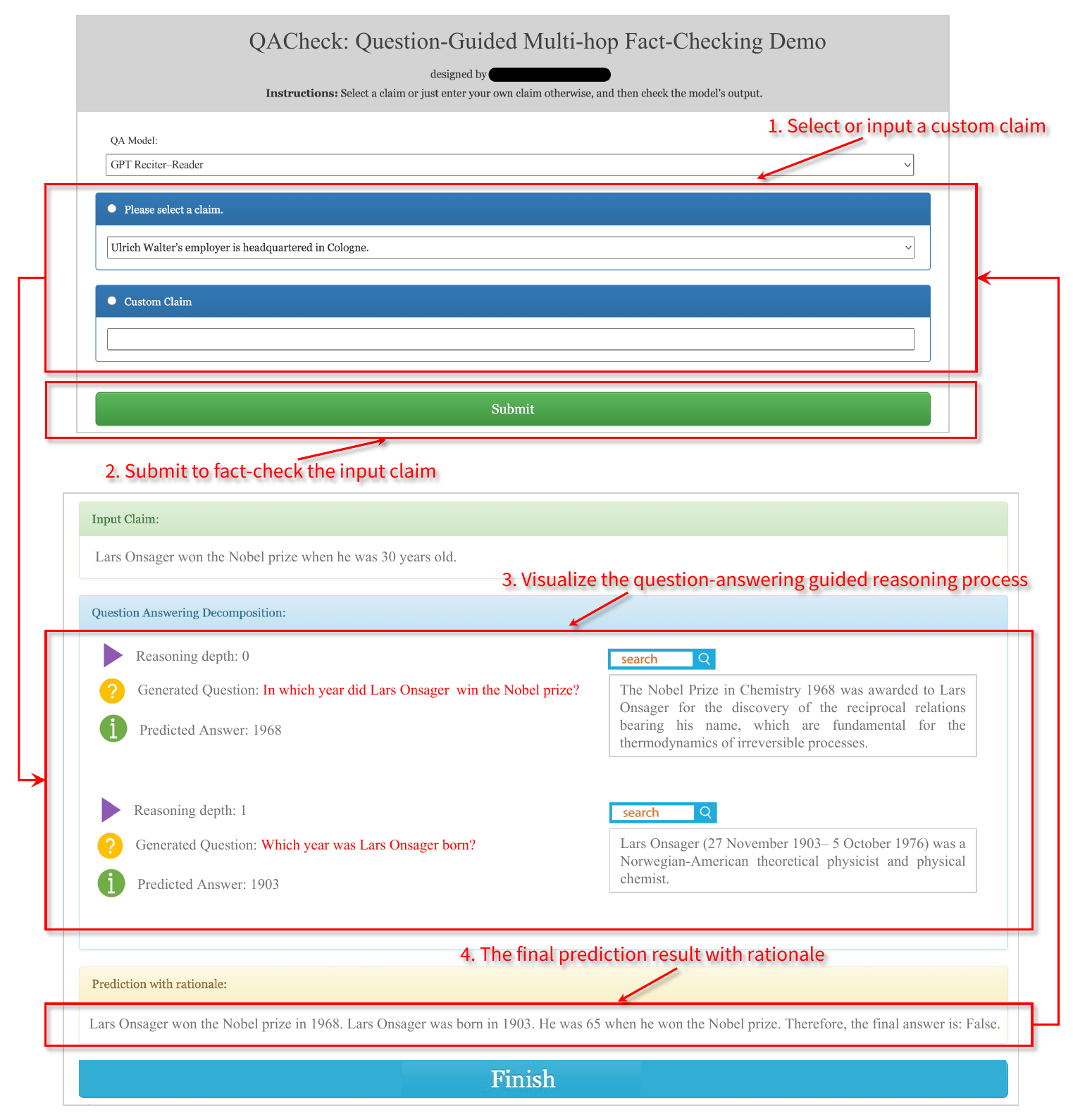}
    \caption{The screenshot of the \system user interface showing its key annotated functions. First, users have the option to \textit{select a claim} or \textit{manually input} a claim that requires verification. Second, users can start the verification process by clicking the \textit{Submit} button. Third, the system shows a step-by-step question-answering guided reasoning process. Each step includes the \textit{reasoning depth}, the \textit{generated question}, relevant retrieved \textit{evidence}, and the corresponding predicted \textit{answer}. Finally, it presents the final prediction \textit{label} with the supporting \textit{rationale}. }
    \label{fig:interface}
\end{figure*}

\subsection{QA Validator}
\label{sec: validator}
The validator module ensures the usefulness of the newly-generated QA pairs. For a QA pair to be valid, it must satisfy two criteria: 1) it brings additional information to the current context $\mathcal{C}$, and 2) it is useful for verifying the original claim. We again implement the validator by prompting InstructGPT with a suite of ten demonstrations shown in Appendix~\ref{subsec:validator}. The instruction is as follows: 
\begin{quote}
\texttt{Claim = \colorbox{gray}{\color{white}{CLAIM}} \\
We already know the following:} \\
\texttt{\colorbox{gray}{\color{white}{CONTEXT}} \\
Now we further know:} \\
\texttt{\colorbox{gray}{\color{white}{NEW QA PAIR}} \\
Does the QA pair have additional \\
knowledge useful for verifying \\ 
the claim?}
\end{quote}
\noindent The validator acts as a safeguard against the system producing redundant or irrelevant QA pairs. Upon validation of a QA pair, it is added to the current context $\mathcal{C}$. Subsequently, the system initiates another cycle of calling the claim verifier, question generator, question answering, and validation.


\subsection{Reasoner}
\label{sec: reasoner}
The reasoner is called when the claim verifier determines that the context $\mathcal{C}$ is sufficient to verify the claim or the system hits the maximum allowed iterations, set to 5 by default. The reasoner is a special question-answering model which takes the context $\mathcal{C}$ and the claim $c$ as inputs and then answers the question \textit{``Is the claim true or false?''}. The model is also requested to output the rationale with the prediction. We provide two different implementations for the reasoner: 1) the end-to-end QA model based on FLAN-T5, and 2) the InstructGPT model with the prompts given in Appendix~\ref{subsec:reasoner}. 



\section{Performance Evaluation}
\label{sec:evaluation}

To evaluate the performance of our \system, we use two fact-checking datasets that contain complex claims requiring multi-step reasoning: HOVER~\cite{Jiang2020HoVerAD} and FEVEROUS~\cite{DBLP:conf/nips/AlyGST00CM21}, following the same experimental settings used in~\citet{pan-etal-2023-fact}. HOVER contains 1,126 two-hop claims, 1,835 three-hop claims, and 1,039 four-hop claims, while FEVEROUS has 2,962 multi-hop claims. We compare our method with the baselines of directly applying InstructGPT with two different prompting methods: (\emph{i})~\textit{direct} prompting with the claim, and (\emph{ii})~\texttt{CoT}~\cite{DBLP:CoT} or chain-of-thought prompting with few-shot demonstrations of reasoning explanations. We also compare with \texttt{ProgramFC}~\cite{pan-etal-2023-fact}, \texttt{FLAN-T5}~\cite{DBLP:journals/corr/abs-2210-11416}, and \texttt{Codex}~\cite{DBLP:Codex}. We use the reported results for the baseline models from~\citet{pan-etal-2023-fact}. 

The evaluation results are shown in Table~\ref{tbl:closed_book_results}. Our \system system achieves a macro-F1 score of 55.67, 54.67, and 52.35 on HOVER two-hop, three-hop, and four-hop claims, respectively. It achieves a 59.47 F1 score on FEVEROUS. These scores are better than directly using InstructGPT, Codex, or FLAN-T5. They are also on par with the systems that apply claim decomposition strategies, \textit{i.e.}, \texttt{CoT}, and \texttt{ProgramFC}. The results demonstrate the effectiveness of our \system system. Especially, the \system has better improvement over the end-to-end models on claims with high reasoning depth. This indicates that decomposing a complex claim into simpler steps with question-guided reasoning can facilitate more accurate reasoning. 

\begin{table}[!t]
\centering
\resizebox{0.48\textwidth}{!}{
\renewcommand{\arraystretch}{1.1}
\begin{tabular}{l|cccc}
\toprule
\multirow{2}{*}{Model} & \multicolumn{3}{c}{HOVER} & \multirow{2}{*}{FEVEROUS} \\
 & 2-hop & 3-hop & 4-hop & \\ \midrule
\texttt{InstructGPT} & & & & \\
\quad \texttt{- Direct} & 56.51 & 51.75 & 49.68 & 60.13 \\
\quad \texttt{- CoT} & \textbf{57.20} & 53.66 & 51.83 & \textbf{61.05} \\ 
\texttt{Codex} & 55.57 & 53.42 & 45.59 & 57.85 \\
\texttt{FLAN-T5} & 48.27 & 52.11 & 51.13 & 55.16 \\ 
\texttt{ProgramFC} & 54.27 & 54.18 & \textbf{52.88} & 59.66 \\ \midrule
\texttt{\system} & 55.67 & \textbf{54.67} & 52.35 & 59.47 \\ 
\bottomrule
\end{tabular}%
}
\caption{Evaluation of F1 scores for different models. The bold text shows the best results for each setting.}
\label{tbl:closed_book_results}
\vspace{-0.1cm}
\end{table}

\section{User Interface}
\label{sec:interface}
We create a demo system based on Flask\footnote{\url{https://flask.palletsprojects.com/en/2.3.x/}} for verifying open-domain claims with \system, as shown in Figure~\ref{fig:interface}. The \system demo is designed to be intuitive and user-friendly, enabling users to input any claim or select from a list of pre-defined claims (top half of Figure~\ref{fig:interface}). 

Upon selecting or inputting a claim, the user can start the fact-checking process by clicking the \textit{``Submit''} button. The bottom half of Figure~\ref{fig:interface} shows a snapshot of \system's output for the input claim \textit{``Lars Onsager won the Nobel prize when he was 30 years old''}. The system visualizes the detailed question-guided reasoning process. For each reasoning step, the system shows the index of the reasoning step, the generated question, and the predicted answer to the question. The retrieved evidence to support the answer is shown on the right for each step. The system then shows the final veracity prediction for the original claim accompanied by a comprehensive rationale in the \textit{``Prediction with rationale''} section. This step-by-step illustration not only enhances the understanding of our system's fact-checking process but also offers transparency to its functioning. 


\system also allows users to change the underlying question--answering model. As shown at the top of Figure~\ref{fig:interface}, users can select between the three different QA models introduced in Section~\ref{sec: qa model}, depending on their specific requirements or preferences. Our demo system will be open-sourced under the Apache-2.0 license. 


\section{Conclusion and Future Works}
\label{sec:conclusion}

This paper presents the \system system, a novel approach designed for verifying real-world complex claims. \system conducts the reasoning process with the guidance of asking and answering a series of questions and answers. Specifically, \system iteratively generates contextually relevant questions, retrieves and validates answers, judges the sufficiency of the context information, and finally, reasons out the claim's truth value based on the accumulated knowledge. \system leverages a wide range of techniques, such as in-context learning, document retrieval, and question-answering, to ensure a precise, transparent, explainable, and user-friendly fact-checking process. 



 

In the future, we plan to enhance \system 1) by integrating additional knowledge bases to further improve the breadth and depth of information accessible to the system~\cite{feng2023factkb,kim2023factkg}, and 2) by incorporating a multi-modal interface to support image~\cite{chakraborty2023factify3m}, table~\cite{chen2020tabfact,lu2023scitab}, and chart-based fact-checking~\cite{akhtar2023reading}, which can broaden the system's utility in processing and analyzing different forms of data. 

\section*{Limitations}



We identify two main limitations of \system. First, several modules of our \system currently utilize external API-based large language models, such as InstructGPT. This reliance on external APIs tends to prolong the response time of our system. As a remedy, we are considering the integration of open-source, locally-run large language models like LLaMA~\cite{DBLP:LLAMA}. Secondly, the current scope of our \system is confined to evaluating \textit{True/False} claims. Recognizing the significance of also addressing \textit{Not Enough Information} claims, we plan to devise strategies to incorporate these in upcoming versions of the system. 

\section*{Ethics Statement}
The use of large language models requires a significant amount of energy for computation for training, which contributes to global warming. Our work performs few-shot in-context learning instead of training models from scratch, so the energy footprint of our work is less. The large language model (InstructGPT) whose API we use for inference consumes significant energy. 
\section*{Acknowledgement}
This project is supported by the Ministry of Education, Singapore, under its MOE AcRF TIER 3 Grant (MOE-MOET32022-0001). The computational work for this article was partially performed on resources of the National Supercomputing Centre, Singapore.

\bibliography{custom.bib}
\bibliographystyle{acl_natbib}

\newpage
\appendix

\section{Prompts}
\label{appen:prompts}
\subsection{Prompts for Claim Verifier}
\label{subsec:claim_verifier}
\lstset{
    style=mystyle,
    basicstyle=\ttfamily\scriptsize,
    backgroundcolor=\color{white},
    stringstyle=\color{black},
    keywordstyle=\color{black},
    breaklines=false,
    keepspaces=false
}

\begin{lstlisting}[language=Python]
(*@\color{codepurple}{\textbf{Claim}}@*) = Superdrag and Collective Soul are 
both rock bands.
We already know the following:
(*@\color{codepurple}{\textbf{Question 1}}@*) = Is Superdrag a rock band?
(*@\color{codepurple}{\textbf{Answer 1}}@*) = Yes
Can we know whether the claim is
true or false now? Yes or no?
(*@\color{codepurple}{\textbf{Prediction}}@*) = No, we cannot know. 

(*@\color{codepurple}{\textbf{Claim}}@*) = Superdrag and Collective Soul are 
both rock bands.
We already know the following:
(*@\color{codepurple}{\textbf{Question 1}}@*) = Is Superdrag a rock band?
(*@\color{codepurple}{\textbf{Answer 1}}@*) = Yes
(*@\color{codepurple}{\textbf{Question 2}}@*)  = Is Collective Soul a rock band?
(*@\color{codepurple}{\textbf{Answer 2}}@*)  = Yes
Can we know whether the claim is
true or false now? Yes or no?
(*@\color{codepurple}{\textbf{Prediction}}@*) = Yes, we can know.

<10 demonstrations in total>
--------
(*@\color{codepurple}{\textbf{Claim}}@*) = [[CLAIM]]
Claim = CLAIM
We already know the following:
[[QA_CONTEXTS]]
Can we know whether the claim is
true or false now? Yes or no?
(*@\color{codepurple}{\textbf{Prediction}}@*) = 
\end{lstlisting}

\subsection{Prompts for Question Generation}
\label{subsec:question_generation}

\textbf{Prompts for the initial question generation}

\lstset{
    style=mystyle,
    basicstyle=\ttfamily\scriptsize,
    backgroundcolor=\color{white},
    stringstyle=\color{black},
    keywordstyle=\color{black},
    breaklines=false,
    keepspaces=false
}

\begin{lstlisting}[language=Python]
(*@\color{codepurple}{\textbf{Claim}}@*) = Superdrag and Collective Soul are 
both rock bands.
To verify the above claim, we can
first ask a simple question:
(*@\color{codepurple}{\textbf{Question}}@*) = Is Superdrag a rock band?

<10 demonstrations in total>
--------
(*@\color{codepurple}{\textbf{Claim}}@*) = [[CLAIM]]
To verify the above claim, we can
first ask a simple question:
(*@\color{codepurple}{\textbf{Question}}@*) = 
\end{lstlisting}

\noindent \textbf{Prompts for the follow-up question generation}

\lstset{
    style=mystyle,
    basicstyle=\ttfamily\scriptsize,
    backgroundcolor=\color{white},
    stringstyle=\color{black},
    keywordstyle=\color{black},
    breaklines=false,
    keepspaces=false
}

\begin{lstlisting}[language=Python]
(*@\color{codepurple}{\textbf{Claim}}@*) = Superdrag and Collective Soul are 
both rock bands.
We already know the following:
(*@\color{codepurple}{\textbf{Question 1}}@*) = Is Superdrag a rock band?
(*@\color{codepurple}{\textbf{Answer 1}}@*) = Yes
To verify the claim, what is the
next question we need to know the
answer to?
(*@\color{codepurple}{\textbf{Question 2}}@*) = Is Collective Soul a rock band?

<10 demonstrations in total>
--------
(*@\color{codepurple}{\textbf{Claim}}@*) = [[CLAIM]]
We already know the following:
[[QA_CONTEXTS]]
To verify the claim, what is the
next question we need to know the
answer to? 
(*@\color{codepurple}{\textbf{Question}}@*) [[Q_INDEX]] = 
\end{lstlisting}

\subsection{Prompts for Validator}
\label{subsec:validator}
\lstset{
    style=mystyle,
    basicstyle=\ttfamily\scriptsize,
    backgroundcolor=\color{white},
    stringstyle=\color{black},
    keywordstyle=\color{black},
    breaklines=false,
    keepspaces=false
}

\begin{lstlisting}[language=Python]
(*@\color{codepurple}{\textbf{Claim}}@*) = Superdrag and Collective Soul are 
both rock bands.
We already know the following:
(*@\color{codepurple}{\textbf{Question}}@*) = Is Superdrag a rock band?
(*@\color{codepurple}{\textbf{Answer}}@*) = Yes
Now we further know:
(*@\color{codepurple}{\textbf{Question}}@*) = Is Collective Soul a rock band?
(*@\color{codepurple}{\textbf{Answer}}@*) = Yes
Does the QA pair have additional
knowledge useful for verifying the claim?
The answer: Yes

<10 demonstrations in total>
--------
(*@\color{codepurple}{\textbf{Claim}}@*) = [[CLAIM]]
We already know the following:
[[QA_CONTEXTS]]
Now we further know:
[[NEW_QA_PAIR]]
Does the QA pair have additional
knowledge useful for verifying the claim?
The answer: 
\end{lstlisting}

\subsection{Prompts for Reasoner}
\label{subsec:reasoner}
\lstset{
    style=mystyle,
    basicstyle=\ttfamily\scriptsize,
    backgroundcolor=\color{white},
    stringstyle=\color{black},
    keywordstyle=\color{black},
    breaklines=false,
    keepspaces=false
}

\begin{lstlisting}[language=Python]
(*@\color{codepurple}{\textbf{Contexts:}}@*)
Q1: When Lars Onsager won the Nobel Prize?
A1: 1968
Q2: When was Lars Onsager born?
A2: 1903
(*@\color{codepurple}{\textbf{Claim}}@*) = Lars Onsager won the Nobel Prize
when he was 30 years old. 

Is this claim true or false?

(*@\color{codepurple}{\textbf{Answer}}@*):
Lars Onsager won the Nobel Prize in 1968. 
Lars Onsager was born in 1903. 
Therefore, the final answer is: False.


<10 demonstrations in total>
--------
(*@\color{codepurple}{\textbf{Contexts:}}@*)
[[CONTEXTS]]
(*@\color{codepurple}{\textbf{Claim}}@*) = [[CLAIM]]
Is this claim true or false?
(*@\color{codepurple}{\textbf{Answer}}@*):
Therefore, the final answer is
\end{lstlisting}

\end{document}